\def\year{2022}\relax
\documentclass[letterpaper]{article} 
\usepackage{aaai22}  
\usepackage{times}  
\usepackage{helvet}  
\usepackage{courier}  
\usepackage[hyphens]{url}  
\usepackage{graphicx} 
\usepackage{natbib}  
\usepackage{caption} 
\DeclareCaptionStyle{ruled}{labelfont=normalfont,labelsep=colon,strut=off} 
\usepackage{algorithm}
\usepackage{algorithmic}

\usepackage{newfloat}
\usepackage{listings}
\floatstyle{ruled}
\newfloat{listing}{tb}{lst}{}
\floatname{listing}{Listing}
\copyrighttext{Presented at the AI-HRI Symposium at AAAI Fall Symposium Series (FSS) 2022}
\title{Considerations for Task Allocation in Human-Robot Teams}

\author{
    Arsha Ali, Dawn M. Tilbury, Lionel P. Robert Jr.
    \\
}
\affiliations{
    University of Michigan \\
    Ann Arbor, MI 48109 \\
    \{arshaali, tilbury, lprobert\}@umich.edu

}

\begin{document}

\maketitle

\begin{abstract}
In human-robot teams where agents collaborate together, there needs to be a clear allocation of tasks to agents. Task allocation can aid in achieving the presumed benefits of human-robot teams, such as improved team performance. Many task allocation methods have been proposed that include factors such as agent capability, availability, workload, fatigue, and task and domain-specific parameters. In this paper, selected work on task allocation is reviewed. In addition, some areas for continued and further consideration in task allocation are discussed. These areas include level of collaboration, novel tasks, unknown and dynamic agent capabilities, negotiation and fairness, and ethics. Where applicable, we also mention some of our work on task allocation. Through continued efforts and considerations in task allocation, human-robot teaming can be improved.

\end{abstract}

\section{Introduction}

The motivation for human-robot teaming is based on the premise that humans and robots are heterogeneous and can have complementary skills. Having complementary strengths and weaknesses can allow humans and robots to work together effectively.

With advanced robots capable of executing certain tasks on their own, one question that arises is how tasks should be allocated in human-robot teams to achieve effective teamwork and performance. Many task allocation methods have been proposed for human-robot teams so the division of tasks is clear, but there are still some areas for continued and further consideration in task allocation methods. This paper presents some of these areas and briefly discusses some of our work on task allocation where applicable. To facilitate our discussion, we refer to an indivisible task simply as a task, where a task is generally allocated to one agent (either a human or a robot).

\section{Background}

An early report, commonly referred to as Fitts' list, details the strengths of humans versus machines. In Fitts' list, the strengths attributed to humans include detection, perception, judgement, induction, improvisation, and long-term memory, while the strengths attributed to machines include speed, power, computation, replication, simultaneous operations, and short-term memory \cite{fitts1951human}. These descriptions have been used as a basis for function allocation \cite{hancock1996future}.  

Following the idea that agents have different capabilities, many human-robot and multi-robot task allocation methods are based on the capabilities of agents and the capabilities required for tasks. Ranz, Hummel, and Sihn's method allocates tasks by matching agent capabilities with task requirements, where agent capabilities consider the elements of cost, time, and quality \cite{ranz2017capability}. Budinsk{\'a} and Havl{\'\i}k's method allocates spatially distributed tasks by considering the binary capabilities and location required for the task with the capabilities and location of robots \cite{budinska2016task}. In addition, AL-Buraiki and Payeur's method assigns robots with specialized capabilities to a task based on the fit between the task and agent and the agent's availability \cite{alburaiki_specialized}.

Task allocation methods have also considered minimizing or moderating factors such as human workload and fatigue. For example, Ge et al. developed an allocation method based on Markov decision process (MDP) that aims to minimize the mental load of all operators in a human-machine system \cite{ge2018mdp}. Frame, Boydstun, and Lopez created a method that continuously evaluates task performance and human workload to redistribute tasks between a human and automation for a surveillance task \cite{frame_allocation}. In addition, Hu and Chen include human fatigue as a continuous-time MDP in their task allocation method \cite{hu_fatigue_allocation}.

Task allocation methods may also be tailored to specific types of tasks or domains, such as assembly tasks. For example, Malik and Bilberg's method allocates assembly tasks between a human and a robot based on the physical features of components (e.g., size, weight, shape), as well as other attributes such as part presentation and safety considerations \cite{malik2019complexity}.

\section{Areas for Consideration}

As described above, there has been great progress and many methods proposed regarding task allocation. In this paper, we identify some areas that could benefit from continued and further consideration in task allocation methods for human-robot teams.

\subsection{Level of Collaboration}

Just as humans and automation can collaborate at different levels (e.g., see \cite{sheridan1978human, parasuraman2000model}), humans and robots can collaborate at different levels too. One study proposes four levels of collaboration including no coexistence (humans and robots are physically separated), coexistence (humans and robots share at least some workspace but do not share goals), cooperation (humans and robots share at least some workspace and have a shared goal), and collaboration (humans and robots share the workspace while working simultaneously on a shared object) \cite{aaltonen2018refining}. Task allocation methods may need to consider the level of collaboration of the human-robot team for a given task for optimal task allocation. Similarly, the type of task allocation problem needs to be well-defined in order to develop an appropriate task allocation method. The level of collaboration may influence the type of task allocation problem. Considering questions such as what constitutes an indivisible task, how many tasks an agent can execute simultaneously, how many agents can be allocated to the same task, what tasks are likely to occur in the future, whether tasks can be re-allocated, and how tasks and agents depend on one another can aid in defining the problem and developing a corresponding task allocation method. 

For example, consider a task of lifting a 50 lbs item with a team of one human and one robot. Consider that the human can lift up to 20 lbs and the robot can lift up to 40 lbs. If there is no coexistence between the human and the robot, simple logic based on capabilities only would indicate that the task of lifting the 50 lbs item should be allocated to the agent that is most capable of the task, or to no agent if no agent on the team is capable. In this example, the task could either be allocated to the robot or the task could be ignored, neither of which is ideal. If the task is allocated to the robot, there is a risk of damage to the item and to the robot. If the task is ignored, future tasks that build on having lifted the 50 lbs item may not be possible. However, if the human-robot team is at the higher levels of collaboration, the agents could lift the 50 lbs item together. The joint collaboration and performance from these two agents is better than one agent acting alone. In a shared workspace with a shared goal, the agents can collaborate simultaneously on the same object. Thus, the level of collaboration and type of problem can influence the outcome of the task allocation method. Fortunately, taxonomies for multi-robot task allocation do exist (e.g., see \cite{gerkey2004formal, korsah2013comprehensive}). These can be used as a starting point to guide the development of human-robot task allocation methods.

\subsection{Novel Tasks}

Novel tasks that the human-robot team has not experienced before may occur, especially in dynamic situations. For example, this could include office, military, and household settings. Thus, task allocation methods would become more robust if they are able to effectively allocate both existing and novel tasks. As a simple example, consider a human-robot team that has previously always been tasked with picking up and sorting cubes and rectangular prisms. In the future, if the tasks also involve picking up and sorting different objects such as spheres and cones, how should these novel tasks be allocated? As a baseline, a task allocation method may default the allocation of all novel tasks to a subset of agents on the team, randomly amongst all agents on the team, or neglect the tasks altogether. However, this may be a suboptimal allocation of novel tasks and better approaches and solutions may exist. 

Part of the challenge in allocating novel tasks has to do with the difficulty in representing and characterizing tasks. Tkach and Amador recently developed a task allocation method for police officers dealing with tasks with unknown locations, arrival times, and importance levels \cite{tkach2021towards}. In our work, we have proposed representing both existing and novel tasks on continuous scales. Tasks are characterized by the levels of different capabilities required for the task \cite{mavric_TAshort, ali2022heterogeneous}. Such a representation allows for a task allocation method to handle both existing and novel tasks, since all tasks are represented in the same manner. However, how to transform a concrete task from the physical environment to a set of required capabilities is not addressed fully in many existing task allocation methods. Knowing the correct levels of capabilities to represent a task is a limitation of our work.

\subsection{Unknown and Dynamic Agent Capabilities}

Agents on a human-robot team may be unfamiliar with the capabilities of their teammates if they have had limited interaction. The question that follows is on how a task allocation method can learn agent capabilities when they are initially unknown. Learning an agent's capabilities is especially applicable to human-robot teams where agents are being swapped in or out (e.g., a robot malfunctions and is replaced by a different robot, humans change during a shift). 

A human can estimate the capabilities of another agent through interactions. Consider a scenario where a human initially thinks another agent can lift 50 lbs. By observing the agent attempting to lift various items, the human updates their estimate of how much that agent can lift. Say the human observed this agent successfully lift 75 lbs, but struggled with lifting 80 lbs. The human can use these observations to approximate the weight the agent can lift to 75 lbs. This updated information can then be used for more informed decision making in the future. When a lifting task arrives, the human has a better idea of how likely it is their teammate could successfully lift the item. If a 25 lbs item arrives, the human would likely feel the agent could lift this item successfully. However, if a 100 lbs item arrives, the human would likely think the agent is incapable of lifting this item successfully. In our work, we use a similar idea for learning an agent's capabilities when they are initially unknown by observing the agent's task performance. We propose representing an agent's unknown capabilities by a belief distribution that gets refined and updated with the history of task outcomes from that agent. The belief in an agent's capabilities informs how much that agent can be trusted to succeed at a task. Trust in an agent then informs the task allocation method \cite{mavric_TAshort, ali2022heterogeneous}.

A related question is how to deal with dynamic agent capabilities. Capabilities may grow through practice or training. Capabilities may also diminish if they are used infrequently or with fatigue. For example, with practice, an agent can improve their capabilities to parallel park a vehicle. However, if these capabilities are not used often, the agent's capabilities to parallel park may be degraded. Observing many task failures or reduced performance from an agent could indicate their capabilities have decreased and need to be reassessed for optimal task allocation. At the expense of failing at a task, an agent could be ``tested" to see if they can succeed on a task that was initially thought to be beyond the agent's capabilities. Again, there may be better approaches and solutions than these preliminary ideas. In short, determining when and how to (re)assess an agent's capabilities is important to avoid poor task allocations.

\subsection{Negotiation and Fairness}

During task allocation, for their own reasons, an agent may disagree with the agent responsible for task allocation. When such disagreements occur, agents will need a way to negotiate the allocation of a task until they reach a consensus. To start, the task allocation method will need to determine whether there are any disagreements among agents. One idea could be to simply request input when an agent disagrees with the allocation of a task. Once it is determined that disagreements between agents are present, how agents will negotiate and whether one agent will have the ultimate authority will have to be considered. 

Roncone, Mangin, and Scassellati propose a negotiation step in a method in which the robot tells a human that it will perform a task or asks the human to perform a task, to which the human can respond ``yes'' or ``no'' \cite{roncone2017transparent}. In the event that two robots intend to work on the same task, Budinsk{\'a} and Havl{\'\i}k prescribe each robot with either a selfish, altruistic, or neutral behavior. The two robots negotiate the task by comparing their behavior type, analogous to a ``rock-paper-scissors'' game \cite{budinska2016task}. Regardless of how negotiation is done, at least for human-robot teams, some form of bi-directional communication will likely be necessary.

Along the same lines, task allocation methods may also need to consider the fairness of resulting task allocations. For example, it may be a poor choice to allocate more tasks or the same types of tasks to one agent compared to other agents. Consider an example in a hospital setting where one agent always gets tasked with preparing deliveries, another agent always makes the deliveries, and another agent always checks in on the patients. On any given day, the frequency of deliveries to prepare and make may be disproportionate to the frequency with which patients need to be checked on. Not only could this mean some agents are overly busy while others are idle, but agents, especially humans, may become tired from doing the same types of tasks. While a robot may not ``care" about the number and types of tasks it is allocated, the allocation of tasks to a robot impacts the allocation of tasks to a human on a human-robot team. A robot could be the agent who typically makes hospital deliveries, but a human may also want to make deliveries to use other capabilities and for ``a change of pace." 

Likewise, prioritizing the preferences of one agent over another in tasks they would like to execute and those they would like to avoid could be perceived as unfair. Unfairness could lead to frustration or low satisfaction in agents, which could in turn harm team performance. One study showed that when a robot distributed wooden blocks unequally between two humans in a tower construction task, team members reported a more negative perception of team relationship compared to teams in which the robot distributed blocks equally \cite{jung_tower}. Going back to the example, a frustrated agent may incorrectly prepare a hospital delivery, which could have serious consequences. Considering a robot as a moral regulator, Kim and Phillips hypothesize that increasing fairness can increase robot legitimacy and the willingness to accept and comply with the robot \cite{kim2021humans}. This hypothesis may extend to a task allocating robot as well. Hence, both negotiation and fairness could help establish and maintain team relationship and team performance.

\subsection{Ethics}

Many task allocation methods are developed such that humans only need to execute tasks as opposed to both allocating and executing tasks, for reasons such as a reduction in human workload and faster completion time. Essentially, a robot or automation serves as the task allocator. In AL-Buraiki and Payeur's method, however, a human operator is involved by selecting a minimum fitting threshold which must be achieved for an agent to be assigned to a task \cite{alburaiki_specialized}. 

An ethical question is whether humans should be allowed to be removed entirely from the role of the task allocator and in what contexts. Even a well-designed task allocation method could inappropriately allocate a task at times (e.g., if the task is a novel task), whereas a human may quickly determine a more appropriate allocation through common sense. In such a case, a human’s input could prevent disastrous outcomes. These issues speak directly to how much autonomy should be given to a robot and the inability to hold a robot accountable \cite{robert2020designing}. Ultimately, in areas where human lives are at risk (e.g., healthcare settings, military settings), task allocation must not only consider who can do what ability-wise but also consider who ethically should be allowed to do what. Embedding ethics into technology is not easy \cite{paul2022intelligence}. Future research must seek to understand the inherent appropriateness and limitations of any task allocation method. Furthermore, if humans are removed from the task allocation process or infrequently negotiate, they may come to rely on and overtrust the task allocation method. If the robot or automation doing the task allocation suddenly becomes unavailable, the human may be ill-prepared to inherit the role of the task allocator. 

These ideas are similar to the role of a semi-autonomous vehicle. A semi-autonomous vehicle may make an inappropriate decision whereas a human may make a better decision using their common sense. Even if temporarily, drivers are removed from the driving task when the vehicle is driving itself and can overtrust the semi-autonomous vehicle in contexts where trust is not justified \cite{mavric_calibrate_trust}. Further, several studies have shown that humans may not be well prepared to resume the driving task if automation becomes unavailable (e.g., see \cite{gold2013take, mok2015emergency,  du2020evaluating}). Semi-autonomous vehicles are designed to operate in some specific contexts, meaning there are times when the human is allowed to disengage to a degree from driving and times when the human needs to drive. Similarly, a task allocation method may at times benefit from human input.

Trust in the task allocation method and the task allocator agent is important. Regarding automation, when there is undertrust, trust is below capabilities and automation is disused \cite{lee2004trust}. When there is overtrust, trust exceeds capabilities and automation is misused \cite{lee2004trust}. Both disuse and misuse can harm performance \cite{mavric_calibrate_trust}. Calibrated trust is when trust matches the automation's capabilities \cite{lee1994trust, muir1987trust}. Similarly, in task allocation, if the task allocation method and task allocator agent are undertrusted, they may not be used and the potential benefits (e.g., reduced workload) cannot be realized. If the task allocation method and task allocator agent are overtrusted, resulting task allocations may be followed in cases that are not justified. Calibrated trust would mean having an appropriate level of trust in the task allocation method and task allocator agent for given purposes and contexts. In other words, there would be an understanding of when to use and when not to use the task allocation method and task allocator agent.

If novel tasks are not a concern and the human-robot team is operating in a well-defined and structured environment, it may be plausible to allow a robot to be the task allocator, relieving cognitive effort from a human. However, if the human-robot team is operating in a dynamic environment where tasks are frequently novel or have serious consequences, it may be more appropriate for a robot to recommend and/or explain task allocations to a human or to have a human serve as the sole task allocator. In sum, the decision on whether humans should be involved in task allocation, to what degree, and how frequently should be considered carefully.

\section{Conclusion}

We present some areas for continued and further consideration in task allocation methods that could improve human-robot teaming and team performance. Depending on the context and what aspects are to be prioritized in a task allocation method, a human may need to be involved in the task allocation decision at some level and work closely with a robot to achieve the best possible allocation of tasks. By improving human-robot teaming, the benefits of introducing robots to society can be realized.

\section{Acknowledgements}

Research was sponsored by the Army Research Office and was accomplished under Cooperative Agreement Number W911NF-21-2-0168. The views and conclusions contained in this document are those of the authors and should not be interpreted as representing the official policies, either expressed or implied, of the Army Research Office or the U.S. Government. The U.S. Government is authorized to reproduce and distribute reprints for Government purposes notwithstanding any copyright notation herein.

\bibliography{ref.bib}
\end{document}